\documentclass[letterpaper]{article} 
\usepackage{aaai25}  
\usepackage{times}  
\usepackage{helvet}  
\usepackage{courier}  
\usepackage[hyphens]{url}  
\usepackage{graphicx} 
\usepackage{natbib}  
\usepackage{caption} 
\usepackage{algorithm}
\usepackage{algorithmic}
\usepackage{amsmath}
\usepackage{booktabs} 
\usepackage{amsfonts}

\usepackage{newfloat}
\usepackage{listings}
\DeclareCaptionStyle{ruled}{labelfont=normalfont,labelsep=colon,strut=off} 
\floatstyle{ruled}
\newfloat{listing}{tb}{lst}{}
\floatname{listing}{Listing}
\title{Capturing the Unseen: Vision-Free Facial Motion Capture \\ Using Inertial Measurement Units}
\author{
    Youjia Wang$^{1, 2*}$, Yiwen Wu$^{1, 2*}$, Hengan Zhou$^{1, 2}$, Hongyang Lin$^{1, 3}$ , Xingyue Peng$^{1}$, Jingyan Zhang$^{1, 4}$, Yingsheng Zhu$^{1}$, Yingwenqi Jiang$^{1}$, Yatu Zhang$^{1}$, Lan Xu$^{1}$, Jingya Wang$^{1}$, Jingyi Yu$^{1}$
    {\tt\small \{wangyj2, wuyw2023, zhouha, linhy, pengxy2023, zhangjy7,} \\
    {\tt\small zhuysh, jiangywq, zhangyt2023, xulan1, wangjingya, yujingyi\}@shanghaitech.edu.cn} \\
}

\affiliations{
    \textsuperscript{\rm 1} ShanghaiTech University \textsuperscript{\rm 2} LumiAni Technology\\
    \textsuperscript{\rm 3} Deemos Technology
    \textsuperscript{\rm 4} ElanTech Co., Ltd.\\
}

\usepackage{bibentry}

\usepackage{color}

\begin{document}

\maketitle

\begin{abstract}
We present Capturing the Unseen (CAPUS), a novel facial motion capture (MoCap) technique that operates without visual signals. CAPUS leverages miniaturized Inertial Measurement Units (IMUs) as a new sensing modality for facial motion capture. While IMUs have become essential in full-body MoCap for their portability and independence from environmental conditions, their application in facial MoCap remains underexplored. We address this by customizing micro-IMUs, small enough to be placed on the face, and strategically positioning them in alignment with key facial muscles to capture expression dynamics.
CAPUS introduces the first facial IMU dataset, encompassing both IMU and visual signals from participants engaged in diverse activities such as multilingual speech, facial expressions, and emotionally intoned auditions. We train a Transformer Diffusion-based neural network to infer Blendshape parameters directly from IMU data. Our experimental results demonstrate that CAPUS reliably captures facial motion in conditions where visual-based methods struggle, including facial occlusions, rapid movements, and low-light environments. Additionally, by eliminating the need for visual inputs, CAPUS offers enhanced privacy protection, making it a robust solution for vision-free facial MoCap.

\end{abstract}

\section{Introduction}

\begin{figure}[t]
    \centering
    \includegraphics[width=1.0\linewidth]{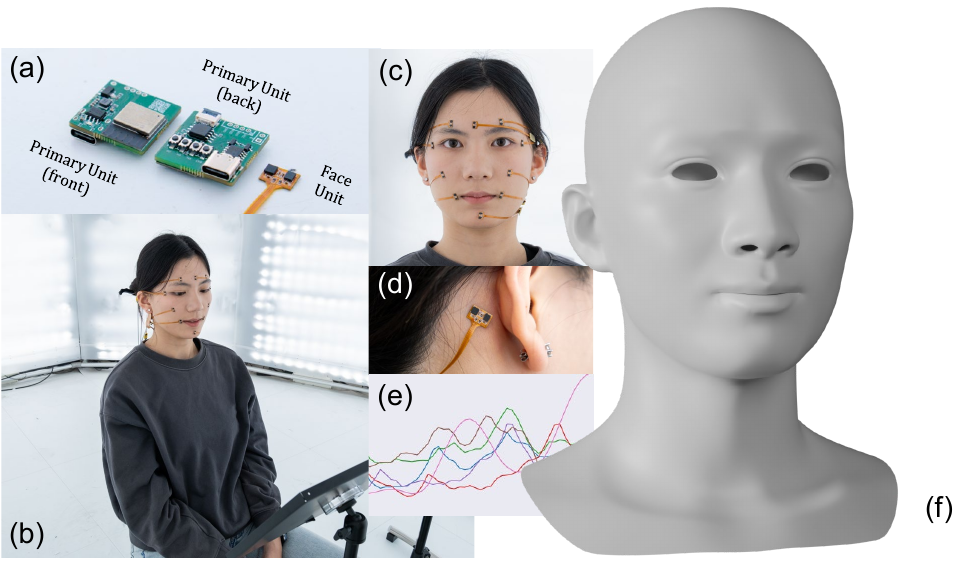}
    \caption{
    We introduce CAPUS, an innovative facial capture system based on IMUs. Using flexible electronic materials, we fabricate miniature IMUs that attach to the human face. Without relying on any visual signals, CAPUS can accurately reconstruct facial expressions.}
    \label{fig:method_data_aqcuisition}
\end{figure}

In the data-driven AI era, the efficacy of analytics tools is directly tied to their ability to adapt to the unique characteristics of different data modalities. For facial motion capture, the success of now widely adopted tools such as 3DDFA \cite{zhu2017face,guo2020towards}, DECA \cite{feng2021learning} and Apple’s ARKit \cite{apple2023} are largely attributed to the availability of RGB and RGBD cameras by mobile devices. Using the captured images as the sole modality, these solutions offer a rapid means of acquiring facial geometry and expression to support various downstream tasks such as editing, relighting, animation, etc \cite{zhang2022videodriven}. However, image as a modality has its own limitations. For example, mobile phone camera based solutions require the user facing the camera all the time, which is impractical in outdoor activities. Existing algorithms are also vulnerable to occlusions, motion blurs, and noise. In fact, for privacy protection, the use of images may even be deliberately avoided.    


In this work, we explore using a new type of data modality for facial motion capture. We observe full-body motion capture using visual signals encounters similar challenges but the latest successes unanimously resorted to the Inertial Measurement Units (IMUs) as the input signal \citet{loper2015smpl}. By attaching the IMUs to various body joints, these solutions manage to capture essential acceleration and axis angle data for modeling body motion. \citet{yi2021transpose} manages to achieve comprehensive body motion capture using as few as six IMUs whereas \citet{li2023ego} leverages the stability and generative capabilities of Transformer Diffusion to further improve the robustness. It is not an exaggeration that IMUs have now become as integral as visual-based methods owing to their exceptional portability and minimal spatial demands. In particular, IMUs neither require using visual sensors nor rely on external environmental conditions, offering unique advantages in outdoor activities and remote applications. 

We introduce Capturing the Unseen (CAPUS), the first IMU-based facial motion capture solution that provides a camera-free alternative to traditional visual-based methods. CAPUS overcomes previous challenges related to the large size and lack of flexibility of IMUs, making them suitable for facial applications. To address this, we developed a new IMU design tailored specifically for facial use as shown in Fig.~\ref{fig:method_data_aqcuisition}(a), with a strong focus on miniaturization. By separating the data acquisition and main control modules, CAPUS ensures that the face-attached device is both compact and lightweight. The acquisition module is designed using flexible materials to adhere comfortably to the face, ensuring accurate signal capture without compromising user comfort. This design minimizes interference with natural facial movements while enabling reliable data transmission and synchronization.


In terms of data processing, we observe that IMU signals tend to have much lower signal-to-noise ratios compared to visual input, leading to less reliable spatial features. Additionally, facial expressions are primarily driven by muscle movements, unlike body motion capture where spatial positions are closely linked to joint rotations. This poses a challenge in effectively interpreting IMU data for facial expressions. To address this, the proposed CAPUS adopts an anatomy-driven strategy by strategically placing IMUs in alignment with specific muscles that control facial expressions. Using CAPUS, we have created the first facial IMU dataset, which includes IMU signals, visual data, and ARKit parameters. This IMU-ARKit dataset records signals from participants performing various activities, such as speaking different languages, making facial expressions, and auditioning with emotional intonation. We then utilize this dataset to train a Transformer Diffusion-based neural network to infer Blendshape parameters directly from the IMU data. Our experiments validate the reliability of the dataset and the effectiveness of our approach.

Moreover, CAPUS supports reliable facial motion capture in traditionally challenging cases for visual-based solutions. In an era where digital privacy is a paramount concern, CAPUS offers a new reliable method of capturing facial expressions without visual input, thereby safeguarding portrait rights. In addition, by freeing a performer from holding a camera by hand toward the face, CAPUS supports facial motion capture while the performer is on the move, with normal body movements to convey body language. Finally, CAPUS can handle challenging scenarios when facial parts (e.g., the mouth) are severely occluded (e.g., during eating or drinking), where vision-based solutions would easily fail. Finally, some subtle changes, especially in the speed of muscle movements, are very challenging to visual sensors but are tractable using IMUs.

In conclusion, our contributions are as follows:

\begin{enumerate}
    \item We introduce the first system capable of recovering human facial expressions using Inertial Measurement Units (IMUs), offering a novel approach to facial motion capture.
    \item We design a new, lightweight IMU device that can be comfortably worn on the face, utilizing flexible electronic materials and weighing just 2.7\% of a commercial Xsens IMU.
    \item A new multi-modal dataset is proposed, which includes aligned IMU signals, visual data, audio signals, ARKit expression parameters, subject emotion labels, and the text of the subject's speech.
    \item We introduce a Transformer Diffusion-based pipeline for inferring Blendshape parameters directly from IMU data, thereby enhancing the capabilities of facial motion capture systems.
\end{enumerate}


\section{Related Works}


\paragraph{Facial Mocap}

Early works by \cite{ferrigno1990pattern,bianchi1998kinematic,guo1994understanding} pioneered the realization of human motion capture. Subsequently, face motion capture systems based on multi-camera setups \cite{michoud2007real,de2004m,vlasic2008articulated,cao2017realtime} became the mainstream solution. Over time, efforts such as \cite{yuan2014localization,von2017sparse} reduced the number of cameras required for effective capture.
During the same period, significant progress was made in facial landmark detection, including both 2D and 3D landmarks \cite{cootes1995active,cootes2001active,cao2014face,zhou2005bayesian}. More recently, specialized 3D reconstruction methods have emerged \cite{bao2021high,smith2020morphable,egger20203d,weise2011realtime,cao2013facewarehouse}, with ARKit being a notable example \cite{apple2023}. However, vision-based approaches are often vulnerable to occlusion issues. The work of \cite{qammaz2023unified} addresses this challenge by predicting information about occluded regions.

\paragraph{Sensor-based Mocap}

The advancements in inertial measurement units (IMUs), driven by works such as \cite{bachmann2001inertial,del2018computationally,foxlin1996inertial,roetenberg2005compensation,vitali2020robust,liu2011realtime,vlasic2007practical,ahmad2013reviews}, have significantly optimized their size and performance, establishing IMUs as a viable tool in the domain of human motion capture.
Early works using IMUs \cite{schepers2018xsens,noitom} achieved full-body human motion capture by mapping IMU rotations to the angles of the human skeleton. Subsequent efforts \cite{huang2018deep,riaz2015motion,slyper2008action,tautges2011motion,von2017sparse,yi2021transpose} have gradually reduced the number of IMUs required for full-body mocap from 17 to as few as 6. These methods offer a broader capture range than vision-based methods and are not constrained by obstacles or lighting conditions.

Some studies \cite{makaussov2020low,mummadi2018real} have utilized IMUs for hand motion capture, demonstrating the potential of IMUs for mocap on smaller body parts.


\section{IMUs for Facial MoCap}

\subsection{Light-weight Facial IMU Sensor Design} 
\begin{figure}[t]
    \centering
    \includegraphics[width=0.9\linewidth]{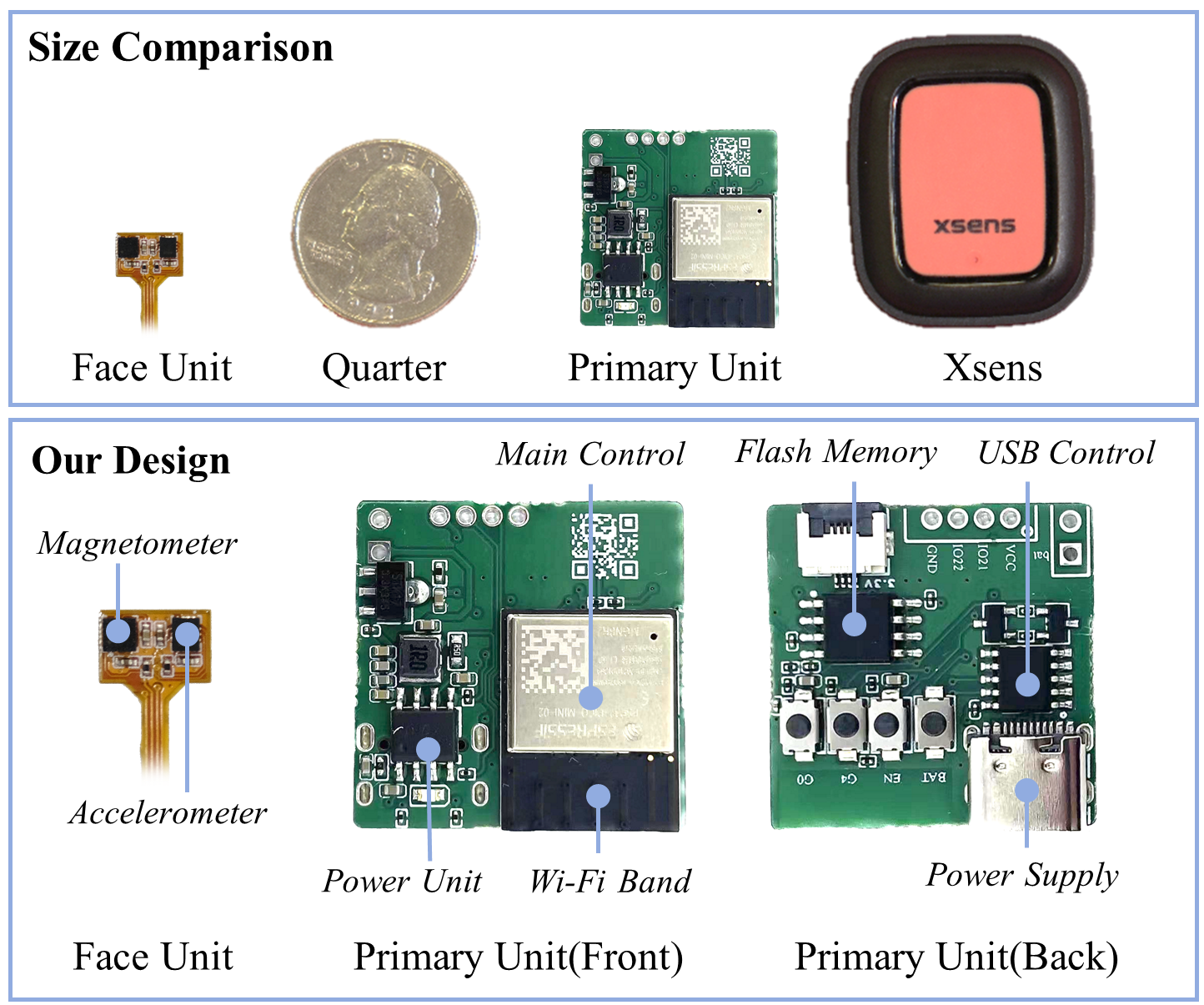}
    \caption{Our IMU has two main components: the face unit and the primary unit. Top: size comparison. Bottom: architecture design. }
    \label{fig:method_imu_design}
\end{figure}

Within the field of motion capture, IMU plays a critical role in reflecting the spatial movements of an object by measuring its orientation and acceleration. IMUs designed for full-body motion capture, such as Xsens, Sony Mocopi, and others, have been widely applied commercially. These units usually consist of various parts, including detecting sensors and data transmission modules, making them too hulking to be used for facial motion capture. Furthermore, employing multiple units of this model for facial capture can lead to severe occlusion, preventing observation of the participant's facial expressions. These necessitates the development of a custom-designed IMU, specifically tailored to meet the unique requirements and scale of facial motion capture.

Our design preserves the function of standard IMU while minimizing weight and size to cater to the requirements for facial capture. 
Fig.~\ref{fig:method_imu_design}(top) compares the size of our IMU. We achieved significant miniaturization by separating the sensor module from the data transmission module. We designed the IMU's face module using flexible electronic materials to closely conform to the skin, ensuring that it does not impede natural facial movements. This design allowed our sensor module to be compact, measuring only 0.6 cm$^2$ and weighing merely 0.3 grams, a stark reduction to 5.4\% the area and only 2.7\% the weight of an Xsens module.

Fig.~\ref{fig:method_imu_design}(down) provides a detailed overview of the specific hardware components utilized in our study. The sensor module incorporates a total of nine-axis sensing sub-units, which include the QMC5883P~\cite{QST} from Silicon Power, a three-axis magnetic field sensor with a measurement range of $\pm 30$ gauss, and the QMI8658~\cite{QST} integrated chip, which combines a three-axis gyroscope and accelerometer. These sensors are capable of accurately recording spatial orientations and accelerations at a rate of 60fps. The data transmission module is primarily based on the ESP32 controller. It employs the UDP protocol to collect and correct data detected by the sensor module. Additionally, we use a Wi-Fi module to transmit the computed data to the host computer. The data includes time stamps, quaternion representations, and acceleration values at each recorded instance.

The data transmission module of our face IMU sensor system requires only a 5V battery supply. This setup provides the essential conditions for the portability and wearability of the face IMU sensor system. Furthermore, the connection to the host computer via Wi-Fi allows users to move freely within the Wi-Fi signal range while wearing the Face IMU, enabling high degrees of mobility.

We further delved deeply into the essential technology for capturing facial information in synchrony using multiple IMUs. To achieve this, it is imperative to address two fundamental challenges: synchronization and calibration.
We designated one ESP32 as the auxiliary ESP32, employing it as a benchmark for synchronizing and calibrating the others. We integrated a calibration program into this ESP32 within the data transfer module during hardware design and used the data module of the auxiliary ESP32's clock as a reference point. We transmitted pulse signals through the DuPont line to each IMU‘s ESP32 for calibration purposes. Upon receiving this pulse signal, each ESP32 aligns its internal clock with the external reference, synchronizing the timestamps across all IMUs. 
Next, acknowledging the variability in facial structures and the potential for slight discrepancies in IMU placement each time, we adopted the concept of a Neutral facial performance, similar to the approach used by ~\cite{yi2021transpose, egger20203d} in body mocap. After wearing the IMUs for the participants, we had each participant relax the facial muscles, presenting a Neutral state, and recorded the orientation of each IMU. In subsequent calculations, we used the orientation relative to this pose as a baseline.
To eliminate the interference with expression prediction caused by head rotation, we strategically place an auxiliary IMU behind the ear, as shown in Fig.~\ref{fig:method_data_aqcuisition}(d). We use this IMU to record the overall rotation of the head. We provide a detailed description in the supplementary materials.

\begin{figure}[t]
    \centering
    \includegraphics[width=0.95\linewidth]{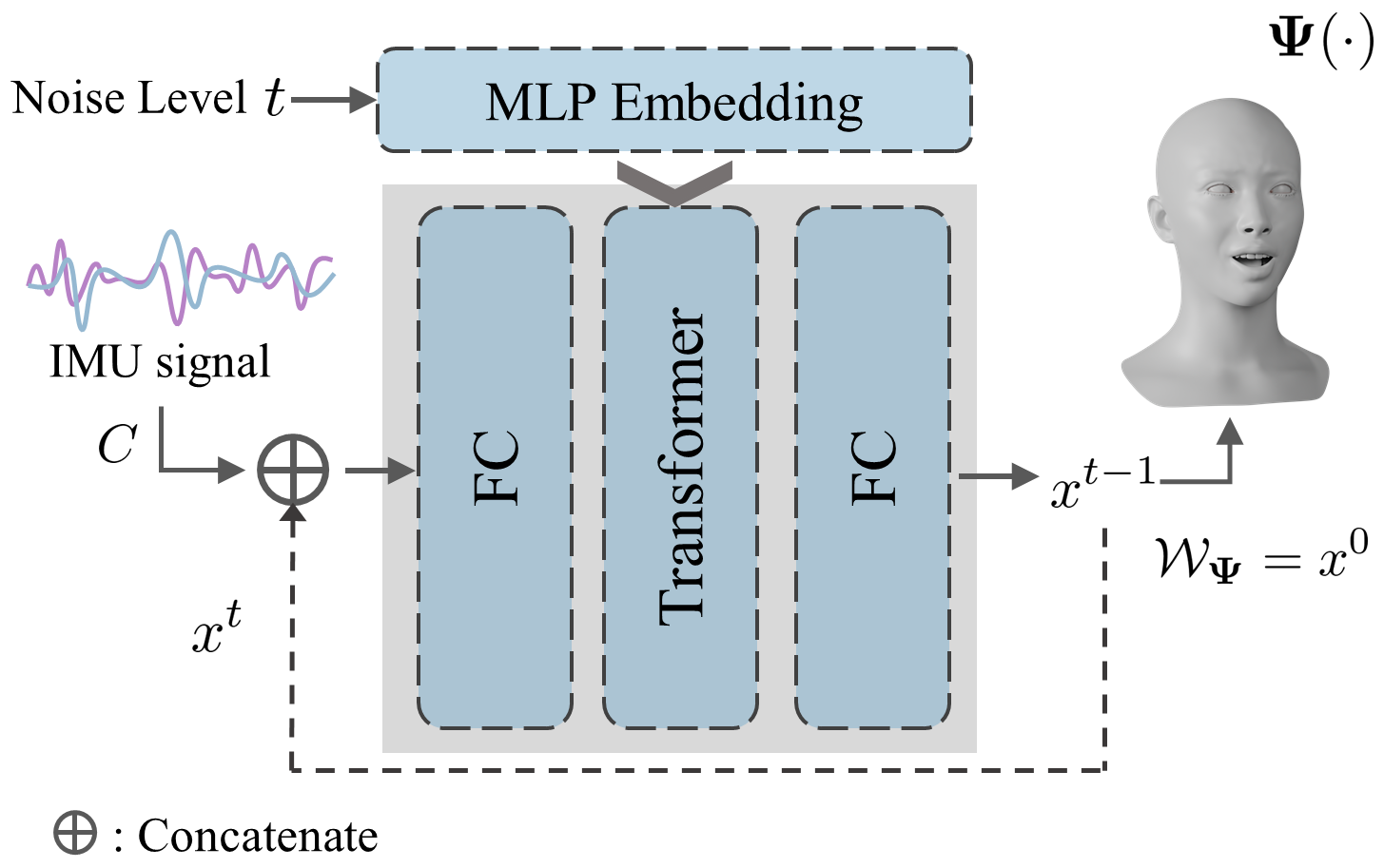}
    \caption{Our transformer diffusion network architecture. We use IMU signal $C$ as a condition input to the network. In each iteration, the network denoises \(x^t\), and finally outputs the predicted blendshape parameters \(x^0\).}
    \label{fig:method_imu_face_model}
\end{figure}


\subsection{Capturing IMU-ARKit Dataset}
\label{location}

To accurately capture facial movements, it is imperative to attach IMUs to distinct regions on the surface of the face. The layout in Fig.~\ref{fig:method_data_aqcuisition}(c) is informed by a detailed analysis of the distribution of facial muscles\cite{Uldis2017}. We demarcated distinct facial zones, the zygomaticus area, the buccinator and mentalis area, the orbicularis oculi area, and the frontalis area. 
In every designated region, we meticulously placed at least one IMU to ensure comprehensive monitoring of the key muscle groups and facial zones.
Acknowledging the sensitivity of certain facial regions, we intentionally avoided placing IMUs on the eyelids and lips, and used the surrounding IMUs to accurately predict their movement.

Our goal is to recover the 3D geometry of the face from the captured IMU data $\mathcal{S}$. A common approach for this task is to use blendshapes as the 3D representation. Blendshape technology, widely adopted in facial animation and motion capture, operates on the principle of parametric modeling, enabling the generation of highly realistic and nuanced expressions. Specifically, a blendshape model is defined by a collection of blendshape weights, denoted as $\mathcal{W} = \{w_1, w_2, \cdots, w_m\}$, 
a facial expression blendshape model can be represented as:

\begin{equation}
\begin{aligned}
M(\mathcal{W}) = B_0 + \sum_k^m w_k B_k.
\end{aligned}
\label{eq:blendshape}
\end{equation}
where $B_0$ represents the neutral face, $B_k$ is the blendshape basis vector, and $m$ is the number of blendshapes. By linearly interpolating between different blendshapes, this approach allows the creation of multiple facial expressions.

Given that the IMU is capable of capturing acceleration and orientation, we propose a method for mapping these physical measurements to blendshape weights $\mathcal{W}$. This requires the development of an algorithm that converts IMU readings into meaningful hybrid shape parameters. 

In order to realize a data-driven solution for predicting facial blendshape weights using IMU, we set out to create a facial IMU dataset aligned with ARKit parameters, as demonstrated in Fig.~\ref{fig:method_data_aqcuisition}. This dataset was carefully compiled to contain paired data of IMU signals and ARKit parameters to ensure a comprehensive base for model training.

Our dataset contains records from 20 different participants. These individuals are all within the 18-40 age range, proficient in English, and have some background in acting, providing richly varied and vivid facial expressions. Fig.~\ref{fig:method_data_aqcuisition}(b) shows an example of the data collection setup. Each participant wore a set of $11$ IMUs and sat in the acquisition seat, with the teleprompter screen placed directly in front of the participant, next to an iPhone that captured the visual information. We used LiveLinkface\cite{llf} to capture the visual information, which is divided into two parts: the RGB video sequence and the ARKit Parameters. 

Before the formal data collection process began, participants were given time to adapt to the sensation of wearing IMUs, ensuring captured facial movements were natural and unrestricted. Participants were instructed to tap the IMU located on mentalis at the start of each recording, as reference frames for synchronizing the IMU signals with the visual signals.
The data was divided into three parts by intentionally designed content that disentangles facial expressions into plain facial movements and emotions. In the first part, participants read aloud the provided content in a calm tone, with a split between native language and English. This was done to capture the natural facial movements associated with the language. In the second part, participants were asked to sequentially make a series of facial expressions that were based on specific classifications, ensuring a full range of emotions and movements. Finally, participants were asked to perform lines of one specific emotion from a set of emotions, joining plain facial movements with emotions.

Our IMU-ARKit dataset provides aligned data pairs of synchronized IMU signals from 11 IMUs, RGB frames, audio signals, and ARKit parameter sequences, with the emotion and content of each sequence annotated. The complete dataset will be accessible for research purposes after acceptance. We showcased samples of our dataset in the supplementary video.

\subsection{IMU-Based Facial Tracker}

Considering the IMU signals provide information not as plain as visual inputs, we chose a lightweight Transformer Diffusion-based network, to interpret the IMU signals meaningfully.

As shown in Fig.~\ref{fig:method_imu_face_model}, our network $\mathbf{\Psi}(\cdot)$ comprises two parts, an MLP embedding network $\mathbf{em}(\cdot)$ and a denoising network $\mathbf{\psi}(\cdot)$. The denoising network has an initial Fully Connected (FC) layer, a concluding FC layer, and a transformer-based core.

A single frame IMU signal $c_j^i = [a_j^i, q_j^i]\in\mathbb{R}^7$ contains acceleration $a_j^i$ and spatial orientation $q_j^i$, where $q_j^i$ is represented as quaternion. We concatenate signals of each frame from all 11 IMUs as $C^i\in\mathbb{R}^{77}$, and stack signals of $T$ consecutive frames to produce the input IMU signal $C\in\mathbb{R}^{T\times 77}$.

\begin{figure}[t]
    \centering
    \includegraphics[width=\linewidth]{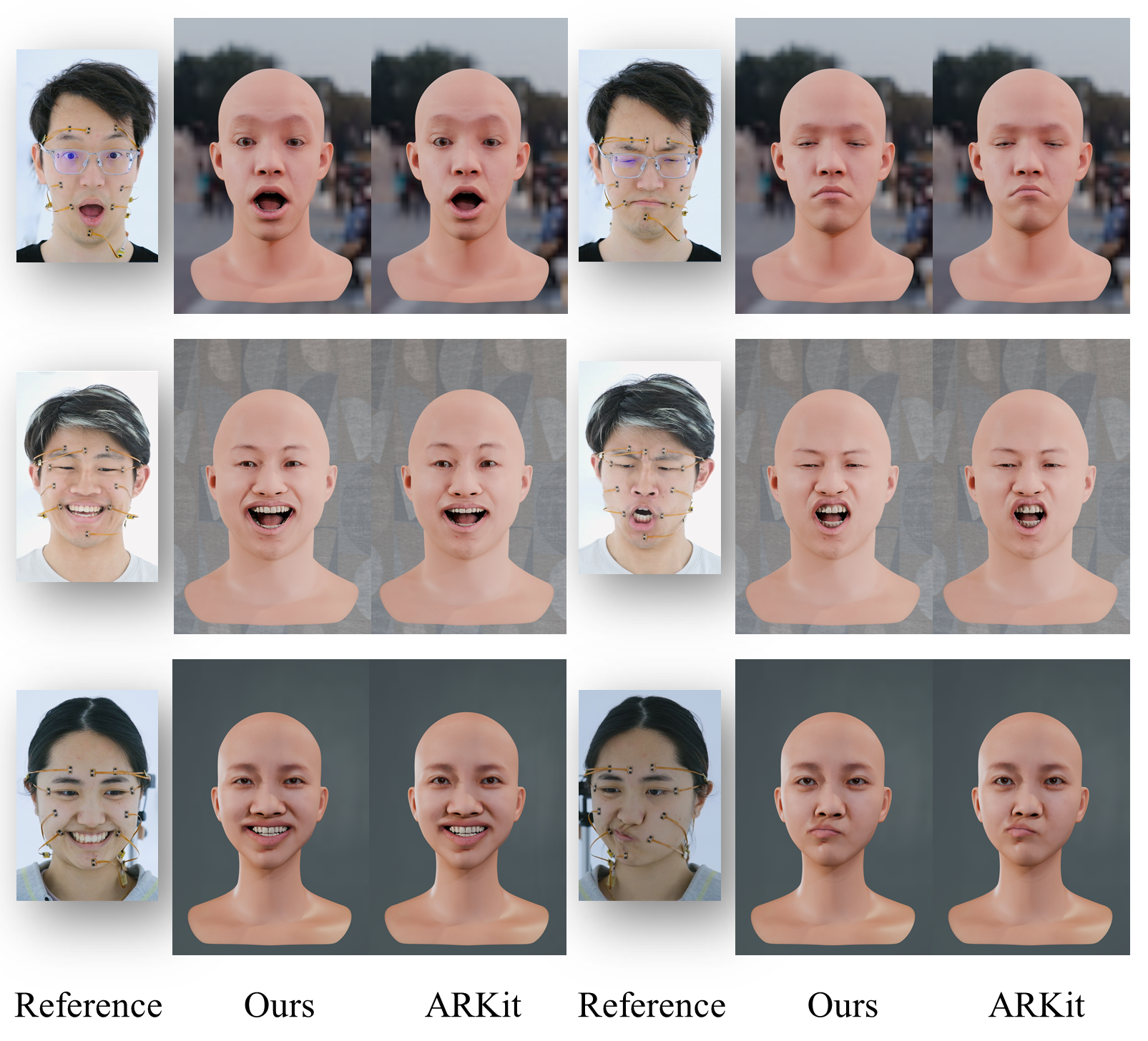}
    \caption{
    Gallery. We present three subjects, with each row corresponding to two different expressions of a single participant. For each subfigure, Left: Image reference. Middle: Facial motion reconstructed by our pipeline. Right: Recorded result by ARKit\cite{apple2023}. Our method achieves results that are comparable to those obtained using ARKit. 
    }
    \label{fig:exp_gallery}
\end{figure}

\begin{figure*}[h]
    \centering
    \includegraphics[width=0.85\linewidth]{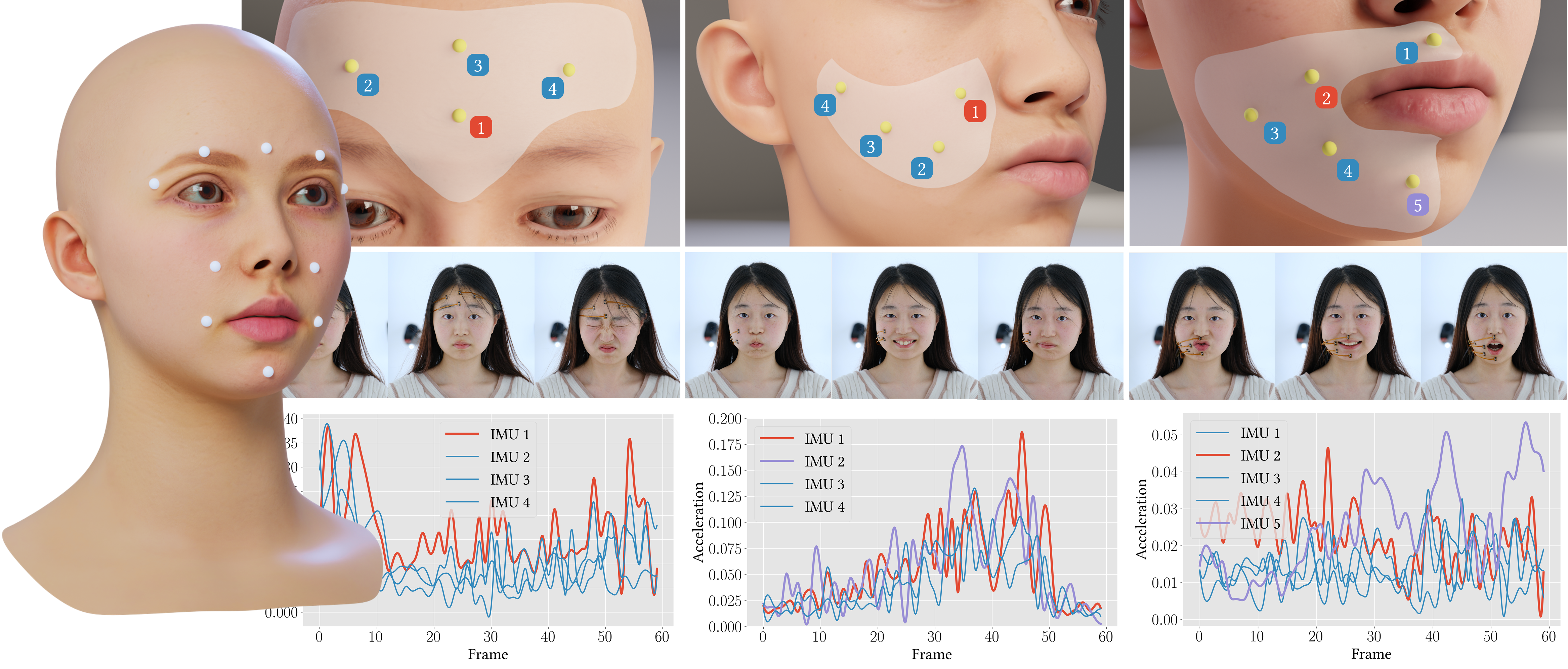}
    \caption{
    Experiment on IMU placement on the face. This figure presents our anatomically-based facial partitioning, highlighting the selected points and the corresponding experiments conducted for each facial region. The left image shows our chosen points on the face, while the other images elaborate on the individual experiments conducted for each specific area. The upper section presents a distribution map of the test points allocated to each region, the middle section identifies the primary expressions and movements associated with that area, and the lower section exhibits the acceleration curves of the IMUs situated at each designated point. 
    }
    \label{fig:exp_imuplacement}
\end{figure*}

\begin{figure*}[h]
    \centering
    \includegraphics[width=1\linewidth]{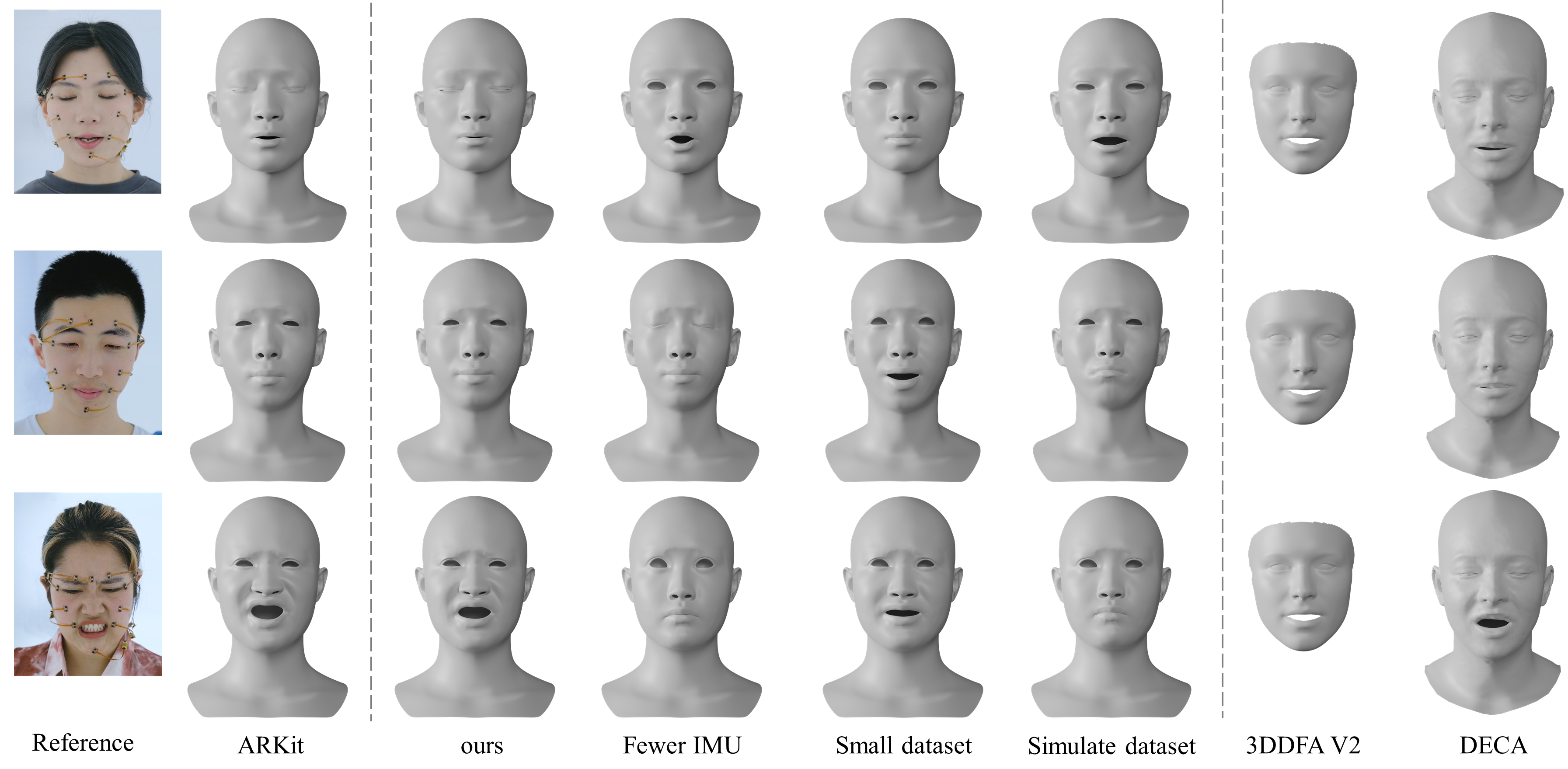}
    \caption{Qualitative comparison and ablation study. The first column displays the reference image. The second column illustrates the record result by ARKit \cite{apple2023}. The third column shows the reconstruction results of our pipeline. Columns 4, 5, and 6 illustrate the result of our ablation experiment \textit{Fewer IMU}, \textit{Small Dataset} and \textit{Simulate Dataset} respectively. Columns 7 and 8 illustrate the results of 3DDFA V2 \cite{guo2020towards} and DECA \cite{feng2021learning} respectively. }
    \label{fig:exp_comparison}
\end{figure*}

To reconstruct blendshape parameters from IMU signals, we utilized the denoising process of the diffusion model with IMU signal $C$ as the condition. Specifically, for the denoising process at noise level $t$, we concatenate the noised blendshape parameters $x^t \in\mathbb{R}^{T\times m}$ with condition $C$, combined with noise embedding $\mathbf{em}(t)$ as input to $\mathbf{\psi}$, and estimate $x^{t-1} \in\mathbb{R}^{T\times m}$. Here $m$ is the number of blendshapes.

\begin{equation}
\begin{aligned}
x^{t-1} = \mathbf{\psi}(\mathbf{em}(t), x^t, C).
\end{aligned}
\label{eq:psi_network}
\end{equation}

We repeat such process until obtaining the predicted blendshape parameters $\mathcal{W}_{\mathbf{\Psi}} = x^0 \in\mathbb{R}^{T\times d}$ as final output.

The paired sequence of blendshape parameters and IMU signals are provided in training. Given ground truth blendshape parameters $\mathcal{W}$, the training loss is defined as follows:

\begin{equation}
\begin{aligned}
\mathcal{L} = \left|| \mathcal{W}_{\mathbf{\Psi}} - \mathcal{W} |\right|_1.
\end{aligned}
\label{eq:loss_fn}
\end{equation}

We trained our network in a supervised manner on our IMU-ARKit dataset, with $T=120$ for both training and testing. To avoid jittering at inference, we set an overlap of $60$ frames to ensure the network has sufficient prior information to accurately determine the initial state of the face within the time window. 

We adopted ICT Face Model \cite{li2020learning} as our blendshapes, and the number of blendshapes $m = 53$. The ARkit parameters are mapped into ICT blendshape parameters.

\section{Experimental Evaluations}

In Fig.~\ref{fig:exp_gallery}, we use CAPUS to recover a variety of facial expressions. We include sequences of facial expressions that represent signature emotions as well as sequences of a performer speaking. The video results can be found in the supplementary video.

Following the similar network architecture as \citet{li2023ego}, CAPUS uses the noise as inputs, imposes the IMU data as transformer conditions, and outputs the inferred blendshape weights to control facial motions. We use Adam as the optimizer with a learning rate $2\times 10^{-4}, \alpha=0.9, \beta=0.999$. We train and evaluate CAPUS on a single NVIDIA RTX3090 GPU. The training process takes $\approx 1$ hours on all identities with paired data. For the generation and rendering of facial assets, we leverage the off-the-shelf technique DreamFace \cite{zhang2023dreamface} to maintain high fidelity and realistic results.

\subsection{Evaluations on IMU Locations}
\label{sec:location}

We qualitatively evaluate how IMU placements across different facial regions affect final facial expression estimation, as shown in Fig.~\ref{fig:exp_imuplacement}. The far left image compares various IMU position schemes, with white dots on the face representing the final locations CAPUS adopts. The images, arranged from left to right, depict the experimental positioning of test points in the Frontalis Area, Zygomaticus Area, and Buccinator and Mentalis Area, respectively. The top row shows the locations we have experimented with for placing the IMUs, with the red and purple ones as the final positions we chose to use. 

In our studies, we strategically select the candidates for placing the IMUs to best reduce interference and align with the underlying muscles. The middle row demonstrates the specific facial movements performed by participants. We collect the acceleration data from respective IMUs during specific facial movements, shown in the bottom row of the images. Our selected IMU locations unanimously produce strong signals that correspond to higher sensitivity under motion. Such placements result in signals with a high SNR suitable for recovering accurate and reliable facial motions. Table ~\ref{tab:diff_region} further shows the quantitative results.

\setlength{\tabcolsep}{1mm}
\begin{table}[t] 

\begin{tabular}{c|c|c|c}
\toprule
 Method  & PVE {[}mm{]}$\downarrow$ & PVE\_LMK {[}mm{]}$\downarrow$ &  MSE$\downarrow$ \\ \midrule
\textbf{Ours}                    & \textbf{0.075 $\pm$ 0.055}      & \textbf{0.126 $\pm$ 0.091}       & \textbf{0.0075}                 \\
\textit{Small Dataset}  & 0.089 $\pm$ 0.063               & 0.150 $\pm$ 0.102                 & 0.0093                           \\
\textit{Fewer IMU}  & 0.078 $\pm$ 0.055               & 0.129 $\pm$ 0.089                 & 0.0082                           \\ \bottomrule
\end{tabular}
\caption{Quantitative ablation study of our method.}
\label{tab:eval_metrics}
\end{table}

\setlength{\tabcolsep}{1.7mm}
\begin{table}[t] 

\begin{tabular}{c|ccccc}
    \toprule
    Area & \#1 & \#2 & \#3 & \#4 & \#5 \hspace{0.3cm} \\
    \midrule
    Frontalis & \textbf{0.71} & 0.57 & 0.52 & 0.58 & - \\
    Zygomaticus & \textbf{0.64} & \textbf{1.20} & 0.57 & 0.32 & - \\
    Buccinator / Mentalis & 0.84 & \textbf{1.38} & 0.54 & 0.88 & \textbf{1.77} \\
    \bottomrule
    
\end{tabular}
\caption{Quantitative evaluations on IMU placements. The table shows the variations times $10^{-3}$ where higher value essentially corresponds to higher sensitivity. Numbers in \textbf{Bold} fonts correspond to the placements that CAPUS uses.}
\label{tab:diff_region}
\end{table}

\subsection{Evaluations on Facial Capture}
\label{sec:imu_face_model_exp}
\label{sec:comp}

Next, we compare CAPUS with the state-of-the-art vision-based techniques DECA \cite{feng2021learning} and 3DDFA\_V2 \cite{guo2020towards}. Specifically, we experiment on a new IMU-ARKit dataset that takes the image captured by iPhone as the input of DECA and 3DDFA\_V2 along with CAPUS. The results are shown in Fig.~\ref{fig:exp_comparison}. 
Columns 3, 7, and 8 correspond to the results from CAPUS vs.  3DDFA\_V2 and DECA. Visual quality wise, CAPUS estimations are comparable to the SOTA visual-based methods. Compared with DECA, CAPUS performs better near the eye region. Compared with 3DDFA\_V2, CAPUS better recovers eyebrow movements induced by facial expressions. 

We then demonstrate the necessity of our dataset. Much work in the field of human motion capture uses simulated datasets for training and tests on a small number of IMU datasets \cite{yi2021transpose, li2023ego}. However, our experiments show that the same approach does not work in facial capture. We used the approach of  \cite{yi2021transpose} to generate a set of simulated datasets using the ARKit parameters of the training set. The performance on the test set after we trained on these simulated data is shown in Figure ~\ref{fig:exp_comparison}. Unlike the performance of human motion capture, the simulated data in face capture does not yield correct results for the network, which fails to make correct predictions in the vast majority of movements. 

We further conduct two ablation experiments to evaluate our dataset and the IMU placements: (1) \textit{Fewer IMU}: we train the network using only a fraction of IMUs, i.e., the ones placed on the eyebrows (2 IMUs), jaw (1 IMU), and cheeks (2 IMUs). (2) \textit{Small Dataset}: we train the network using 1/3 of the dataset. 

The variations are illustrated in columns 4, and 5 of Fig.~\ref{fig:exp_comparison} sequentially. The results in column 4 show some examples that CAPUS fails to faithfully predict the motion, e.g., closed eyes. This is largely attributed to the locations where we place the IMUs. The results in column 5 manage to recover challenging facial distortions under extreme expressions. This indicates that our training dataset is sufficiently rich to cover these movements and the trained network is robust enough to generalize to reproduce these distortions.

\begin{figure}[t]
    \centering
    \includegraphics[width=0.95\linewidth]{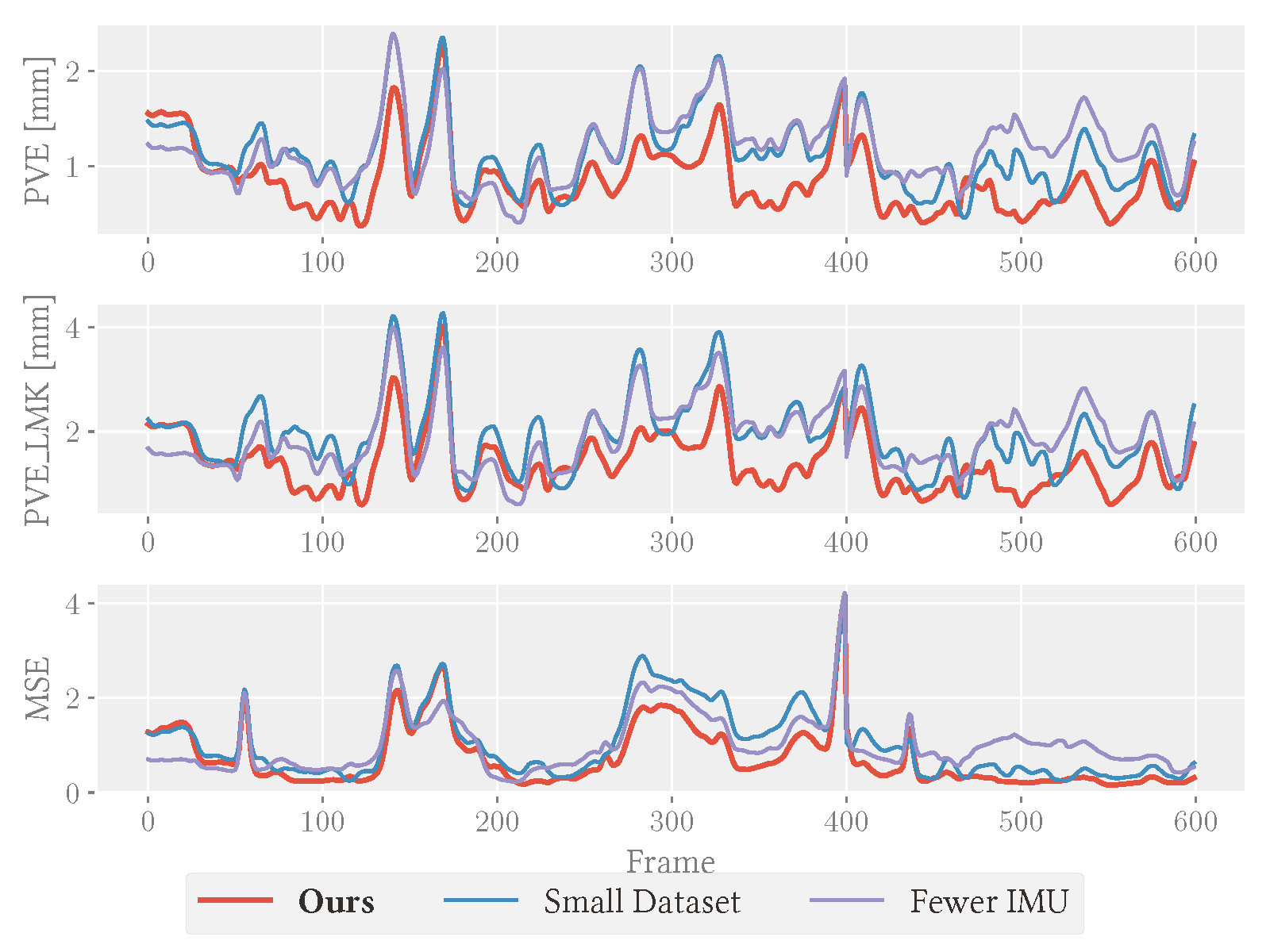}
    \caption{Quantitative result of our method on test data. We plot the PVE, PVE\_LMK and MSE calculated per frame with ARKit as ground truth on a sequence. }
    \label{fig:exp_eval_metrics}
\end{figure}

We further conduct quantitative evaluations in Fig.~\ref{fig:exp_eval_metrics} and Table.~\ref{tab:eval_metrics}. Same as \cite{feng2021learning, guo2020towards}, we calculate the 3D per vertex error (PVE)\cite{decaf} on the deformed mesh as an indicator of the similarity between ARKit vs. IMSUE predictions. Specifically, we use the 3D landmark vertex error (PVE\_LMK) to demonstrate the fidelity of CAPUS estimations on visually significant areas. We further calculate the MSE using the predicted blendshape weights with ARKit as the ground truth. The red curve represents the metrics for each frame using CAPUS whereas the purple and blue curves represent the metrics for the two ablation experiments. 

In the supplementary material, we further compare our network with other architectures, and demonstrate that our network is not overfitted to the training set.

\subsection{Applications}
\paragraph{Camera-Free Facial Capture}

In traditional facial capture systems, users need to always face the camera, which limits the head and body movements. For example, while on the move, users have to hold their phones by hand, making it difficult to perform normal body movements and convey body language.
We demonstrate using CAPUS as a portable facial capture solution, as shown in the supplementary video. 
Due to the modular design of the IMUs, the user's facial skin experiences minimal weight. All IMUs are powered by a portable power bank, using Wi-Fi module to communicate with the computer. As a result, CAPUS allows for accurate facial capture while a person is walking, preserving complete facial information and freeing the user's hands. 

\begin{figure}[h]
    \centering
    \includegraphics[width=0.9\linewidth]{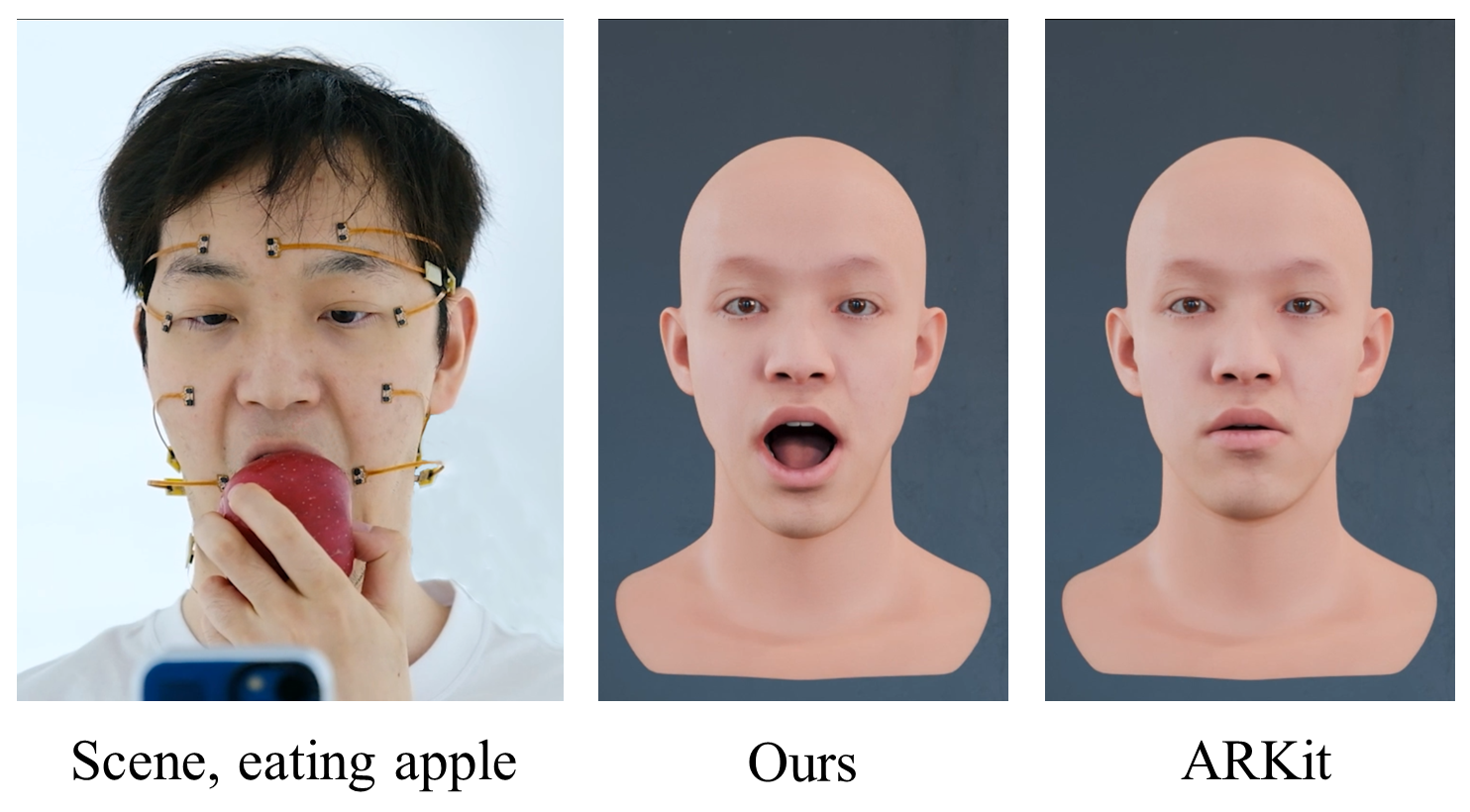}
    \caption{Facial capture during occlusions.}
    \label{fig:app_all}
\end{figure}

\paragraph{Occluded Facial Capture}
In some scenarios, facial capture encounters unavoidable occlusions, such as during eating or drinking. Professional actors commonly resort to 'mimicking' eating to avoid this issue, which can result in a lack of authenticity. 
We demonstrate using CAPUS to conduct robust motion capture in such scenarios. We showcase CAPUS's accurate and stable motion capture capabilities in the heavily occluded 'eating an apple' situation, as shown in Fig.~\ref{fig:app_all}. While eating, the user's hands and the food largely occlude the face, particularly the mouth regions, rendering vision-based methods ineffective. CAPUS, instead, does not rely on any video signals, allowing for accurate capture of the mouth movements. The supplementary video includes several dynamic sequences.

\section{Conclusion and Discussion}

We have presented CAPUS, a novel vision-free facial motion capture technique that takes only IMU signals as input. Our tailored micro-IMUs, strategically attached to facial regions aligned with facial anatomy, enable us to capture facial movements from nuanced to dramatic. We have collected the first-ever IMU-ARKit dataset with synchronized IMU and visual signals of diverse expressions from various performers. We further developed a learning-based framework for reliable motion inference. Both the dataset and the code will be released to the community for comprehensive evaluations.

We believe our IMU-based facial motion capture is an innovative and potentially advantageous solution. In full-body motion capture, due to the exceptional portability and minimal spatial demands of IMUs, we have witnessed a transition from vision-only to vision-IMU hybrid and most recently to IMU-only solutions. Similarly, we believe that IMU-based methods will also become mainstream in the field of facial motion capture. CAPUS illustrates this possibility by achieving results comparable to visual methods while allowing users to move their heads freely. This is particularly advantageous in scenarios where visual signals are unavailable or intentionally avoided, such as industrial facial capture solutions that require subjects to wear helmets, or smartphone-based solutions that require users to always face the camera. As a prototype, CAPUS is far from perfect and still has many issues (comfort, sensor sizes, wiring, etc). Yet, we believe using IMU as a new modality in facial motion capture may stimulate significant future developments in facial animation, capture, and beyond.

\clearpage


\clearpage
\newpage


\bibliography{aaai25}

\end{document}


\maketitle
\section{Overview}
Our supplementary materials include the following:
\begin{enumerate}
    \item Supplementary text (this PDF): including some details added to the Method section, and some experiments on network training.
    \item Supplementary video: including an overall introduction to the paper, and some demonstrations of dynamic results.
    \item Data: due to the large volume of our data, in the system, we only submit some sample data to evaluate the correctness and reproducibility of our algorithm. After the paper is accepted, the complete dataset will be open source.
    \item Code: We have prepared two versions of the code. \texttt{code\_demo.zip} corresponds to the submitted data for evaluating the reproducibility of our algorithms, and the way to run the code is written in the \texttt{readme.md}. \texttt{code.zip} is the code of our complete project. After the paper is accepted, all the code will be collated and open source.
\end{enumerate}

\section{Calibration}

The calibration process of our IMU design consists of two parts: sensor calibration and coordinate calibration. 

The sensor calibration calibrates the magnetic field sensor with an integrated calibration program, ensuring the orientations are calculated accurately. We adopted the same calibration procedure as the Magnetic Field Mapper function \cite{xsens}, calibrating by rotating the IMU sensor in all directions for one minute. During this process, the IMU captures magnetic data $M = [M_x, M_y, M_z]$, filters the outliers, and automatically updates the calculated offset and scale for each axis $k$ as  

\begin{equation}
    offset_k = \frac{\max(M_k) + \min(M_k)}{2}
\label{eq:calib_mag_offset}
\end{equation}
\begin{equation}
    scale_k = \frac{\text{mean}(\max(M) - \min(M))}{\max(M_k) - \min(M_k)}
\label{eq:calib_mag_scale}
\end{equation}

This manual calibration process is required at most once a day for a fixed capture environment. 

The coordinate calibration processes the captured IMU signals to a unified local coordinate space, dealing with the impacts of the variability in facial structures and the potential for slight discrepancies in IMU placement each time. We adopted the concept of a Neutral facial performance, similar to the approach of T-pose used by ~\cite{yi2021transpose, egger20203d} in body MoCap. The Neutral facial performance is defined as the orientation $\mathcal{R}_{neutral}^i$ of each IMU attached to the participant's face surface, with facial muscles relaxed, representing the Neutral state of the participant. For captured IMU signal of $i^\mathrm{th}$ IMU $\mathcal{S}^i = [a_{raw}^i, q_{raw}^i]$ aligned by frame index, with $\mathcal{R}_{raw}^i$ denoting the corresponding rotation matrix of $q_{raw}^i$, we have the relative rotation represented as: 

\begin{equation}
\begin{aligned}
\mathcal{R}_{rel}^i = (\mathcal{R}_{neutral}^i)^{-1}\mathcal{R}_{raw}^i
\end{aligned}
\label{eq:calib_rot_rel}
\end{equation}

Considering the accelerations $a_{raw}^i$ are recorded under local coordinate space, with each IMU's orientation varied, we aligned the accelerations to world coordinate as: 

\begin{equation}
\begin{aligned}
a_{align}^i = (\mathcal{R}_{raw}^i)^{-1}a_{raw}^i
\end{aligned}
\label{eq:calib_acc_align}
\end{equation}

Under capturing situations, two types of movements influence an IMU's signal when placed on the face — the overall movement of the head and the movements caused by facial expressions. Our focus is on deducing expressions from IMU signals. We strategically placed an auxiliary IMU behind the ear at a location specifically designed, and mapped the orientations and accelerations to the local coordinate space of the auxiliary IMU to mitigate the impact of general head movements on the orientations and accelerations detected by the other IMUs. For convenience, we define the index of the auxiliary IMU to be $0$, then the calibrated IMU rotation and acceleration can be expressed as
\begin{equation}
\begin{aligned}
\mathcal{R}_{calib}^{i} = \left\{ 
  \begin{array}{ll}
    (\mathcal{R}_{rel}^{0})^{-1} \mathcal{R}_{rel}^{i}  & \text{if } i\neq 0, \\
    \mathcal{R}_{rel}^{i} & \text{otherwise}
  \end{array} 
\right.
\end{aligned}
\label{eq:calib_rot}
\end{equation}
\begin{equation}
\begin{aligned}
a^{i} = \left\{ 
  \begin{array}{ll}
    (\mathcal{R}_{rel}^{0})^{-1} a_{align}^{i}  & \text{if } i\neq 0, \\
    a_{align}^{i} & \text{otherwise}
  \end{array} 
\right.
\end{aligned}
\label{eq:calib_acc}
\end{equation}

After this automatic process, we obtained the calibrated IMU signal $C^i = [a^i, q^i]$ that is used for training the network, with $q^i$ the corresponding quaternion representation of $\mathcal{R}_{calib}^i$.

\setlength{\tabcolsep}{1.7mm}
\begin{table*}[t] 
    \centering
\begin{tabular}{c|c|c|c}
    \toprule
    Network & RNN & LSTM & Ours \\
    \midrule
    PVE {[}mm{]}$\downarrow$ & $0.073\pm 0.068$ & $0.066\pm 0.072$ & \textbf{0.060$\pm$ 0.069} \\
    PVE\_LMK {[}mm{]}$\downarrow$ & $0.124\pm 0.117$ & $0.112\pm 0.122$ & \textbf{0.101$\pm$ 0.119} \\
    MSE$\downarrow$ & 0.00665 & 0.00648 & \textbf{0.00526} \\
    \bottomrule
\end{tabular}
\caption{Network Architecture Comparison. }
\label{tab:network_comparison}
\end{table*}

\section{Facial-IMU Simulation}

Simulate IMU datasets generated from other types of MoCap data have been widely applied in body MoCap using IMU, as a means to enlarge the training dataset. Benefiting from the SMPL model, synthetic IMU data can accurately reflect the orientation and acceleration of actual IMU attached to corresponding body parts. 

Inspired by this approach, we exploited the possibility of simulating IMU signals on the facial surface. We utilized the basis provided by the ICT Face Model \cite{li2020learning}, using the ARKit sequences to generate a topologically consistent face mesh for each frame. Given the corresponding positions of IMU $i$ on the face mesh at frame $j$ in vertices as $v_j^i \in \mathbb{R}^{3}$, for each IMU $i$, the simulated acceleration is calculated as follows: 
\begin{equation}
\begin{aligned}
a_j^i = \frac{v^i_{j-n} + v^i_{j+n} - 2v^i_j}{n\tau}
\end{aligned}
\label{eq:simulate_acc}
\end{equation}
where $\tau$ is the time interval of two frames, and $n$ is the constant that controls the degree of smoothing.

To determine orientation, we focused on the facial region containing the selected vertex. We select a face that contains the vertex, denoted as $\{v^i_j, v^{i, \text{sup}[1]}_j, v^{i, \text{sup}[2]}_j\}$. We identified two adjacent edges to this vertex, labeled as $e^{i, 1}_j = (v^i_j - v^{i, \text{sup}[1]}_j)_{\text{norm}}$ and $e^{i, 2}_j = (v^i_j - v^{i, \text{sup}[2]}_j)_{\text{norm}}$. The first vector from these edges was utilized as the x-axis for the IMU coordinate system. We calculated the rotation matrix of the IMU relative to the world coordinate system as
\begin{equation}
\mathbf{R}_j^{ i} = 
\begin{bmatrix}
\mathbf{e}_j^{i, 1}, & \frac{\mathbf{e}_j^{i, 1} \times \mathbf{e}_j^{i, 2}}{\|\mathbf{e}_j^{i, 1} \times \mathbf{e}_j^{i, 2}\|}, & \frac{(\mathbf{e}_j^{i, 1} \times \mathbf{e}_j^{i, 2}) \times \mathbf{e}_j^{i, 2}}{\|(\mathbf{e}_j^{i, 1} \times \mathbf{e}_j^{i, 2}) \times \mathbf{e}_j^{i, 2}\|}
\end{bmatrix}
\label{simulate_rot}
\end{equation}

As with the captured data, we calibrate the synthetic data by Eq.~\ref{eq:calib_rot_rel}, Eq.~\ref{eq:calib_rot}, and Eq.~\ref{eq:calib_acc} to obtain the data used for network training.
This comprehensive approach to simulating both the acceleration and orientation of IMUs ensures that our dataset is not only extensive but also rich in the detailed, accurate representation of facial movements.

We synthesized simulate facial IMU data and conducted experiment using this data as training set, with testing result on captured data shown in Fig.6: Qualitative comparison and ablation study. The result indicates the simulation approach for body IMU data is not viable for facial IMU data. 
\section{Supplementary Experiment}
\paragraph{Network Architecture}

For inferring sequential data, the mainstream baseline models include RNN~\cite{zaremba2015recurrentneuralnetworkregularization}, LSTM~\cite{lstm} and DiT~\cite{ho2020denoisingdiffusionprobabilisticmodels}. Given the low-dimensional nature and low SNR of IMU signals, we opted for a DiT-based method, as demonstrated by \citet{li2023ego}. We conduct an evaluation of the network structure with 1) a simple RNN and 2) a modified version of the LSTM network in \citet{yi2021transpose} using the same training and testing settings. Specifically, consistent with our 4-layer DiT network with $d\_model=512$, the RNN contains 4 layers with 512 hidden units each, the LSTM network contains 4 layer with 512 hidden units each. 
We then assessed the different network structures in terms of accuracy. As shown in Table~\ref{tab:network_comparison}, our DiT-based network structure achieved the lowest discrepancy. 

\begin{figure}[t]
    \centering
    \includegraphics[width=1.0\linewidth]{figures/exp_convergence.png}
    \caption{
    Training and Evaluation Loss. }
    \label{fig:exp_convergence}
\end{figure}

\paragraph{Convergence}
For our training strategy, we selected a portion of the data from each participant as the training set, while the remaining data was used as the test set. The test set differs from the training set in both speaking content and the subjects' expressions.  

We analyzed the learning curves by plotting the training and evaluation losses over time. As shown in Fig.~\ref{fig:exp_convergence}, both the training and validation losses converge with similar trends. This indicates that the network has learned the underlying patterns rather than simply memorizing individual fragments. 

\bibliography{aaai25}